%% file: main.tex
\documentclass[conference]{IEEEtran}
\IEEEoverridecommandlockouts

\usepackage{cite}
\usepackage{amsmath,amssymb,amsfonts}
\usepackage{graphicx}
\usepackage{textcomp}
\usepackage{xcolor}
\usepackage[hidelinks]{hyperref}

\def\BibTeX{{\rm B\kern-.05em{\sc i\kern-.025em b}\kern-.08em
    T\kern-.1667em\lower.7ex\hbox{E}\kern-.125emX}}

\begin{document}

\title{SEAT: Sign-Type Evidence-Audited Transport for Multi-Label Tongue Diagnosis%
\thanks{This work was supported in part by the National Natural
Science Foundation of China under Grant Nos. 62603130,
62522309, and 62473088.
\textit{(\mbox{*Corresponding author: Yining Qian; email: qianyining@qhd.neu.edu.cn.})}\par\noindent This work has been submitted to the IEEE for possible publication. Copyright may be transferred without notice, after which this version may no longer be accessible.\par\noindent \textcopyright{} 2026 IEEE. Personal use of this material is permitted. Permission from IEEE must be obtained for all other uses, in any current or future media, including reprinting/republishing this material for advertising or promotional purposes, creating new collective works, for resale or redistribution to servers or lists, or reuse of any copyrighted component of this work in other works.}}

\author{
\IEEEauthorblockN{Liuyi Xu\textsuperscript{1}, Qing Wang\textsuperscript{1},
Zikai Liang\textsuperscript{2}, An-Yang Lu\textsuperscript{1},
Yining Qian\textsuperscript{3,*}}
\IEEEauthorblockA{\textsuperscript{1}College of Information Science and Engineering,\\
Northeastern University, Shenyang, China}
\IEEEauthorblockA{\textsuperscript{2}College of Sciences,
Northeastern University, Shenyang, China}
\IEEEauthorblockA{\textsuperscript{3}School of Computer and Communication Engineering,\\
Northeastern University at Qinhuangdao, Qinhuangdao 066004, Hebei, China}
}

\hypersetup{
  pdftitle={SEAT: Sign-Type Evidence-Audited Transport for Multi-Label Tongue Diagnosis},
  pdfauthor={Liuyi Xu, Qing Wang, Zikai Liang, An-Yang Lu, Yining Qian}
}

\IEEEtriggeratref{5}

\maketitle

\begin{abstract}
Automated tongue-sign recognition can improve the consistency of non-invasive
examinations that otherwise rely on subjective visual judgment. A single tongue image
often exhibits multiple coexisting signs characterized by heterogeneous evidence,
including localized morphology, global chromaticity, and inter-label dependence.
However, existing methods typically rely on shared representations and uniform
multi-label heads, making it difficult to verify whether predictions stem from
clinically relevant evidence rather than spurious visual correlations. We propose SEAT,
an evidence-audited framework that matches distinct evidence types with dedicated
prediction and evaluation mechanisms. Built on frozen DINOv2 patch features, SEAT
couples mask-guided sign-specific attention for localized morphology, a semi-relaxed
optimal transport pathway over LAB/HSV color prototypes to enhance pale-tongue
recognition, and a co-occurrence-aware transport regularizer to shape the batch-level
predicted label marginal. These components are evaluated through audits matched to their
evidence types. On the TongueDx benchmark, SEAT achieves a mean macro-F1 of 72.59 with a
standard deviation of 1.04 across six random seeds, with a six-model ensemble reaching
74.15. The ensemble outperforms a backbone-matched baseline and a larger,
domain-specialized vision-language model by 8.06 and 2.99 percentage points,
respectively. Our results demonstrate that aligning dedicated prediction pathways and
audits with sign-specific characteristics improves multi-label accuracy while offering
more targeted evidential grounding than uniform post-hoc spatial explanations. Code is
available at \url{https://github.com/NEU-TCM-Lab/Sign-type-Evidence-Audited-Transport}.
\end{abstract}


\input{sections/01_introduction}
\input{sections/03_method}
\input{sections/04_experiments}
\input{sections/06_conclusion}

\end{document}

%% file: sections/01_introduction.tex
\section{Introduction}

Tongue inspection is a routine, non-invasive part of clinical assessment, and
automating it could make an otherwise subjective examination more consistent. A
single tongue image may show several signs at once, such as tooth marks, cracks,
red spots, ecchymosis, a thick or yellow coating, and abnormal tongue-body
color, so the task is naturally a multi-label recognition problem rather than a
mutually exclusive classification one~\cite{bib:drtongue}, and, as in other
medical multi-label settings, it exhibits strong label co-occurrence and class
imbalance~\cite{bib:chexpert,bib:cxrlt}. The central difficulty is that these
signs are defined by fundamentally different kinds of visual evidence: tooth
marks and cracks by localized morphology, a pale tongue body by global
chromaticity, and still others by regional color or coating texture~\cite{bib:tcmtongue}. Accurate
multi-label prediction alone therefore does not establish that a model has
relied on evidence consistent with a sign's clinical definition. This kind of
evidence is precisely what clinically interpretable decision support requires.

Modern recognition systems build on deep image encoders and have benefited from
transformer backbones~\cite{bib:vit}, large-scale self-supervised and vision-language
pretraining, and extensions to
multi-label recognition~\cite{bib:mldecoder}. A common multi-label formulation encodes an image with a shared backbone
and predicts all signs through a common head. In broader
multi-label learning, graph~\cite{bib:mlgcn,bib:ssgrl} and label-query~\cite{bib:q2l} architectures model dependencies
among outputs, and post-hoc attribution
visualizes influential regions~\cite{bib:gradcam,bib:rise}, though such spatial
maps themselves require faithfulness
checks~\cite{bib:attention_not_explanation,bib:sanity}. These methods
improve accuracy, dependency modeling, or spatial interpretation, but these
capabilities alone do not establish whether predictions rely on the clinical
cues that should support the corresponding signs.

This gap matters because different evidence types demand different assessments.
For a localized sign such as a tooth mark, spatial reliance can be probed by
suppressing highly weighted patches and measuring the change in the prediction.
For a chromatic sign such as tongue-body paleness, a heatmap can show where the
model attends but cannot establish whether the prediction follows clinically
relevant variation in lightness, redness, or saturation. Label co-occurrence is
different again: it is not a property of any image region but an empirical
output-space structure best examined through the label-space mechanism it
informs. A shared encoder may contain all of these signals implicitly, yet
neither a common head nor a single explanation method~\cite{bib:sanity,bib:onemap} ensures they are modeled
and evaluated according to their distinct roles.

Motivated by this, we propose SEAT (Sign-Type Evidence-Audited Transport), which
pairs each clinically distinct evidence type with a matching prediction
mechanism and evidence audit on frozen DINOv2
features~\cite{bib:dinov2}. Its discriminative core is a mask-guided,
per-sign attention reader that aggregates patches separately for each of
the eight signs. For the chromatic pale-tongue sign, SEAT adds a color-prototype
pathway over patch-aligned LAB/HSV descriptors that recovers fine-grained color
underrepresented by frozen semantic features. For output-level structure, a
co-occurrence-informed regularizer shapes the batch-level predicted label
marginal under a cost geometry derived from training co-occurrence, solved with
entropic optimal transport~\cite{bib:sinkhorn,bib:peyre}, without fixing
per-image label combinations. Each mechanism is audited by a matched procedure:
readout-mask suppression for morphology, color counterfactuals for chromaticity, and
inspection of the learned transport coupling for label structure.

The main contributions of this work are as follows:
\begin{enumerate}
  \item We formulate multi-label tongue-sign recognition as a
  heterogeneous-evidence problem, distinguishing localized morphology and
  tongue-body chromaticity from an output-space prior derived from empirical
  label co-occurrence. This formulation enables evidence-specific modeling
  without assigning an independent pathway to every label.

  \item We propose SEAT, an evidence-audited multi-pathway framework that combines a
  mask-guided sign-specific reader, an explicit color-prototype pathway for
  TonguePale, and a co-occurrence-informed regularizer over the batch-level label
  marginal. These components provide complementary inductive biases for spatial,
  chromatic, and output-space information.

\item We evaluate both predictive performance and evidence response. On
  TongueDx, SEAT achieves a mean macro-F1 of $72.59 \pm 1.04$ across six seeds,
  while the six-model ensemble reaches $74.15$, outperforming both a competitive
  baseline built on the same DINOv2-L backbone with a global head ($66.09$) and a
  recent domain-specialized vision-language model, ShizhenGPT-7B-VL ($71.16$), by
  $8.06$ and $2.99$ points, respectively. The full model improves its underlying
  attention reader by $1.11$ points on average. Patch-suppression and
  color-counterfactual analyses support the intended behavior of the spatial and
  color pathways, while the transport coupling makes the label-space prior
  directly inspectable.
\end{enumerate}

%% file: sections/03_method.tex
\section{Methodology}

\begin{figure*}[t]
\centering
\includegraphics[width=0.98\textwidth]{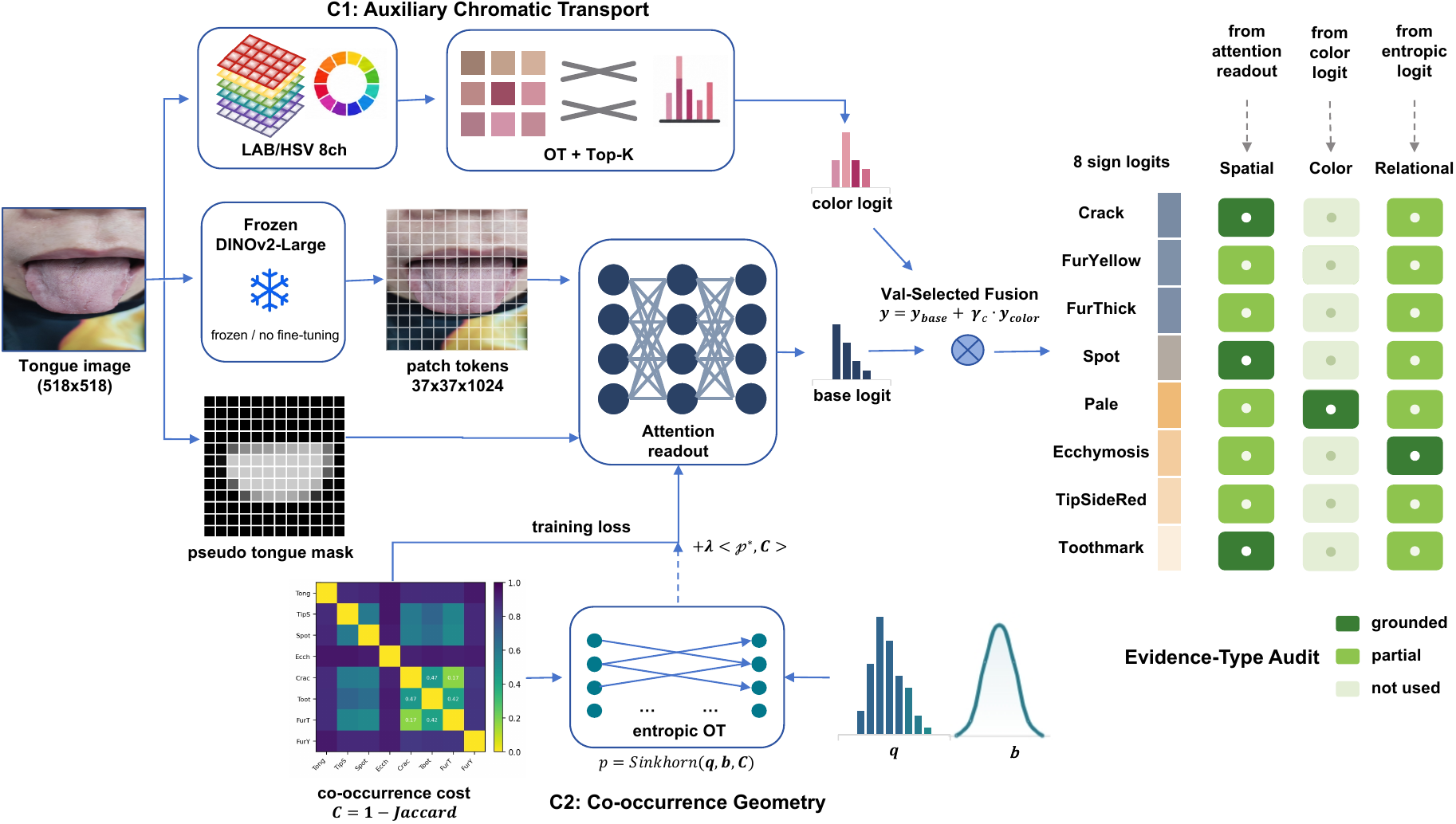}
\caption{Overview of SEAT. A frozen DINOv2-Large backbone encodes each image into a $37\times37$ patch grid. A mask-guided multi-head softmax attention reader forms the discriminative core. C1 adds a color-prototype optimal-transport pathway over per-patch LAB/HSV descriptors, whose pale-tongue output is combined with the reader by a validation-selected late-fusion weight. C2 adds a co-occurrence regularizer that shapes the batch-level predicted label marginal under a $1-\text{Jaccard}$ cost geometry.}
\label{fig:overview}
\end{figure*}

Guided by the principle that each sign should be predicted and audited using the cue that clinically defines it, we define SEAT as follows: a mask-guided attention reader on a frozen backbone, a color pathway for the pale-tongue sign (C1), and a co-occurrence-informed regularizer on the predicted label marginal (C2). The three pathways share one optimal-transport formulation whose $\lambda\!\to\!0$ limit is softmax attention; the discriminative reader is most effective at this attention limit (Section~\ref{sec:exp-ablation}), so transport contributes the shared form and the C1/C2 inductive bias rather than discriminative accuracy. The reader and C1 share a $37\times37$ patch grid. The reader operates on cached DINOv2 tokens, whereas C1 operates on per-patch color descriptors from the same image. C2 acts on the batch-level label marginal produced by the reader. The evidence audit that evaluates these predictions is separate, defined with its results in Section~\ref{sec:experiments}. An overview is shown in Fig.~\ref{fig:overview}.

\subsection{Problem Setup and Frozen Patch Features}

We first fix the notation and the frozen features shared by all three pathways. Each tongue image carries a multi-hot label $y\in\{0,1\}^{C}$ with $C=8$ signs in the fixed order TonguePale, TipSideRed, Spot, Ecchymosis, Crack, Toothmark, FurThick, FurYellow. An image is resized to $518\times518$ and encoded by a frozen DINOv2-Large backbone~\cite{bib:dinov2}, following the patch-token representation popularized by vision transformers~\cite{bib:vit,bib:mae}. Its last hidden state yields $N=37\times37=1369$ patch tokens $X=\{x_n\}_{n=1}^{N}$ with $x_n\in\mathbb{R}^{1024}$. A pseudo tongue mask is average-pooled onto the same grid to give patch weights $m_n\in[0,1]$; it comes from a frozen open-vocabulary segmentation pipeline applied identically to all splits, with no manual annotation (Section~\ref{sec:experiments}). The backbone sees the full image, and the mask only biases the readout. We use the normalized mask prior $\bar{m}_n=(m_n+\delta)/\sum_j(m_j+\delta)$ with $\delta=10^{-4}$ where a patch distribution is required. Frozen DINOv2 patch tokens are cached once and shared across all DINO-token readout variants; C1 additionally uses deterministic color descriptors aligned to the same grid.

\subsection{Mask-Guided Attention Reader}
\label{sec:readout}

The discriminative core is a per-sign multi-head attention reader. Patch tokens are projected as $\tilde{x}_n=W_p\,\mathrm{LN}(x_n)\in\mathbb{R}^{d}$ ($d{=}256$), and split into $H{=}8$ heads of width $d/H$. For head $h$, sign $c$ has a learnable query $q_c^{(h)}$; keys and values are $k_n^{(h)}=W_K^{(h)}\tilde{x}_n$ and $v_n^{(h)}=W_V^{(h)}\tilde{x}_n$. Queries and keys are $\ell_2$-normalized, so attention uses cosine similarity under a mask log-bias and a fixed temperature $\tau=0.1$:
\begin{equation}
A^{(h)}_{c,n}=\mathrm{softmax}_{n}\!\left(\frac{\cos(q_c^{(h)},k_n^{(h)})}{\tau}+\log(m_n+\delta)\right).
\label{eq:attn}
\end{equation}
The head-wise pooled features $z_c^{(h)}=\sum_n A^{(h)}_{c,n}v_n^{(h)}$ are concatenated across heads and read out by a per-sign linear layer,
\begin{equation}
z_c=\big[z_c^{(1)};\dots;z_c^{(H)}\big],\qquad \hat{s}^{\mathrm{base}}_c=w_c^{\top}\mathrm{LN}(z_c)+b_c .
\end{equation}
This reader is the core of SEAT; C1 and C2 add structure around it rather than replacing it.

\subsection{C1: Auxiliary Chromatic Transport with Pale-Tongue Fusion}
\label{sec:color}

The pale-tongue sign is defined by chromaticity rather than shape, a property that frozen semantic features may underrepresent, so we add an explicit color pathway. On the same $37\times37$ grid, each patch $n$ carries an eight-dimensional color descriptor $u_n$: the per-patch means of six channels obtained by OpenCV RGB-to-LAB and RGB-to-HSV conversion scaled to fixed ranges ($L,a,b,H,S,V$), plus the per-patch standard deviations of $L$ and $S$. Statistics are computed over all pixels within a patch, and the mask enters only later as patch weights. We do not apply additional illumination normalization, to preserve clinically meaningful chromatic variation. C1 is a color pathway over these descriptors, built on a mask-guided semi-relaxed optimal transport~\cite{bib:sinkhorn,bib:peyre} and trained with the same weighted binary cross-entropy as the reader. It uses shared color projection layers $(W^{\mathrm c}_p,W^{\mathrm c}_K,W^{\mathrm c}_V)$, together with per-sign prototype banks $\{r_{c,k}\}_{k=1}^{K}$ and per-sign color classifiers. Although the multi-label color loss supervises all signs, only the TonguePale logit, the one sign whose clinical definition is chromatic, is retained for late fusion (Section~\ref{sec:objective}).

For sign $c$ the pathway keeps $K{=}8$ learnable prototypes $\{r_{c,k}\}_{k=1}^{K}$. Let $\tilde{u}_n=W^{\mathrm{c}}_p\mathrm{LN}(u_n)$ be the projected color descriptor, $\phi(u_n)=\overline{W^{\mathrm{c}}_K\tilde{u}_n}$ its $\ell_2$-normalized key, and $\hat{v}_n=W^{\mathrm{c}}_V\tilde{u}_n$ its value. With prototype--patch similarity $\cos(r_{c,k},\phi(u_n))$, the mask enters as an additive log-bias folded into a mask-biased cost,
\begin{equation}
\tilde{C}^{(c)}_{k,n}=\underbrace{1-\cos(r_{c,k},\phi(u_n))}_{\text{color mismatch}}\;-\;\varepsilon_{\mathrm{col}}\log(m_n+\delta),
\end{equation}
so the mask term is active regardless of the relaxation strength. The plan solves a semi-relaxed entropic OT with the prototype marginal fixed uniform ($\mu=\mathbf{1}_K/K$) and the patch marginal KL-relaxed toward the normalized mask prior $\bar{m}$:
\begin{equation}
\begin{split}
P_c^{\star}=\arg\min_{P\ge 0,\ P\mathbf{1}=\mu}\ &\langle P,\tilde{C}^{(c)}\rangle+\varepsilon_{\mathrm{col}}\!\sum_{k,n}P_{k,n}(\log P_{k,n}-1)\\
&+\lambda_{\mathrm{col},c}\,\mathrm{KL}\!\left(P^{\top}\mathbf{1}\,\big\|\,\bar{m}\right),
\end{split}
\label{eq:colorot}
\end{equation}
with entropic temperature $\varepsilon_{\mathrm{col}}=\exp(\eta_{\mathrm{col}})$, a single global learnable scalar (clamped to $[10^{-3},1]$) initialized to $0.1$. C1 uses $20$ unrolled Sinkhorn iterations. Because the mask log-bias sits inside the cost $\tilde{C}^{(c)}$, it stays active even as $\lambda_{\mathrm{col},c}\!\to\!0$. In that limit the KL term vanishes and each normalized prototype row reduces to $\mathrm{softmax}_n\big(\cos(r_{c,k},\phi(u_n))/\varepsilon_{\mathrm{col}}+\log(m_n+\delta)\big)$, recovering the attention form of Eq.~\eqref{eq:attn} in the color-descriptor space. The per-sign relaxation strength $\lambda_{\mathrm{col},c}=\sigma(\rho_c)\in(0,1)$ controls the soft preference for the aggregate patch marginal $\bar{m}$: values near zero recover this unconstrained attention-like limit, whereas larger values couple the prototype assignments more strongly through $\bar{m}$.

The color logit sums a pooled term and a peak term. Each prototype row of $P_c^{\star}$ is normalized to a patch distribution $\bar{a}_{k,n}=P^{\star}_{c,k,n}/\mu_k$ with $\sum_n\bar{a}_{k,n}=1$; the per-prototype pooled vectors $\sum_n\bar{a}_{k,n}\hat{v}_n$ are concatenated over the $K$ prototypes into $e_c$, giving the pooled logit $\hat{s}^{\mathrm{pool}}_c=w^{\mathrm{c}\top}_c\mathrm{LN}(e_c)+b^{\mathrm{c}}_c$ with the color head's own classifier parameters. A peak term then summarizes the strongest color matches. With the mask-biased match score $\xi_{c,n}=\max_k\cos(r_{c,k},\phi(u_n))+\log(m_n+\delta)$ and $\mathcal{T}_c$ the $K_{\mathrm{peak}}{=}16$ patches of largest $\xi_{c,n}$,
\begin{equation}
\pi_c=\frac{1}{K_{\mathrm{peak}}}\sum_{n\in\mathcal{T}_c}\xi_{c,n},\qquad
\hat{s}^{\mathrm{color}}_c=\hat{s}^{\mathrm{pool}}_c+g_c\,\pi_c ,
\end{equation}
where $g_c$ is a learnable per-sign scalar controlling the contribution of the peak term. The mask log-bias steers $\xi_{c,n}$ toward tongue patches while retaining soft support at uncertain boundaries. The pooled term captures tongue-wide chromatic evidence, while the peak term responds to the most color-consistent patches.

\subsection{C2: Co-occurrence Geometry for Label-Marginal Regularization}
\label{sec:coocc}

Tongue signs exhibit empirical co-occurrences: some sign pairs appear together more frequently than others in the training labels. Per-label binary cross-entropy scores each output independently and does not use this structure. We add a co-occurrence regularizer that shapes the batch-level predicted label marginal under a cost geometry derived from these empirical co-occurrence patterns. It does not impose per-image co-occurrence constraints: the quantity it regularizes is a marginal over the $C$ signs, and pairwise structure enters only through the ground cost.

From the training labels we compute the pairwise Jaccard co-occurrence and its induced ground cost,
\begin{equation}
J_{i,j}=\frac{\sum_{r}\mathbf{1}(y_{r,i}\!=\!1\wedge y_{r,j}\!=\!1)}{\sum_{r}\mathbf{1}(y_{r,i}\!=\!1\vee y_{r,j}\!=\!1)},\qquad C_{i,j}=1-J_{i,j},
\end{equation}
so frequently co-occurring signs are close and rarely co-occurring signs are far. For a batch $B$ with $p_{r,c}=\sigma(\hat{s}^{\mathrm{base}}_{r,c})$ and training count $n_c=\sum_{r\in\mathcal{D}_{\mathrm{train}}}y_{r,c}$, the predicted label marginal and the smoothed empirical frequency are
\begin{equation}
\begin{aligned}
q_c&=\frac{\sum_{r\in B}p_{r,c}}{\sum_{j=1}^{C}\sum_{r\in B}p_{r,j}},\\
b_c&=\frac{n_c+\alpha}{\sum_{j=1}^{C}(n_j+\alpha)}
=\frac{n_c+\alpha}{\sum_{j=1}^{C}n_j+C\alpha},
\end{aligned}
\end{equation}
with $\alpha$ a smoothing constant; in implementation we use the equivalent frequency-domain form with $\alpha=10^{-6}$. The normalization makes C2 constrain the relative allocation of predicted positive mass across signs rather than the total predicted positive mass. We run $50$ Sinkhorn iterations to obtain the entropic transport coupling between $q$ and $b$ under cost $C$,
\begin{equation}
\begin{split}
P^{\star}
&=\mathrm{Sinkhorn}_{\varepsilon_{\mathrm{coocc}}}(q,b;C) \\
&\approx \arg\min_{P\in\Pi(q,b)}
\Big[\langle P,C\rangle \\
&\quad +\varepsilon_{\mathrm{coocc}}
\sum_{i,j}P_{i,j}(\log P_{i,j}-1)\Big].
\end{split}
\end{equation}
where $\Pi(q,b)=\{P\ge 0: P\mathbf{1}=q,\ P^{\top}\mathbf{1}=b\}$. The regularizer is the geometry cost induced by this entropy-regularized coupling between the predicted and empirical label marginals:
\begin{equation}
R_{\mathrm{coocc}}(q,b)=\langle P^{\star},C\rangle .
\label{eq:rcoocc}
\end{equation}
The entropic term shapes the coupling $P^{\star}$ but does not itself enter the objective; gradients are propagated through the unrolled Sinkhorn iterations, so $R_{\mathrm{coocc}}$ is a differentiable function of $q$, with $b$ constant. This regularizer leaves the attention reader unchanged and acts only on the aggregate output distribution.

\subsection{Joint Objective and Inference}
\label{sec:objective}

The reader and C2 are trained jointly in one run; the color pathway is trained separately; and the two are combined by validation-selected late fusion. All three use the same weighted binary cross-entropy
\begin{equation}
\begin{split}
\mathcal{L}_{\mathrm{WBCE}}(y,\hat{s})
= -\sum_{c=1}^{C}\Big[
&w^{+}_c\,y_c\log\sigma(\hat{s}_c) \\
&+(1-y_c)\log(1-\sigma(\hat{s}_c))
\Big].
\end{split}
\end{equation}
with per-class positive weight $w^{+}_c=\mathrm{clip}(N^{-}_c/N^{+}_c,\,0.2,\,5.0)$ from the training positive/negative counts. The reader with C2 minimizes
\begin{equation}
\mathcal{L}_{\mathrm{base}}=\mathcal{L}_{\mathrm{WBCE}}\big(y,\hat{s}^{\mathrm{base}}\big)+\lambda_{\mathrm{coocc}}\,R_{\mathrm{coocc}}\big(q(\hat{s}^{\mathrm{base}}),b\big),
\end{equation}
with $\lambda_{\mathrm{coocc}}=0.32$ and $\varepsilon_{\mathrm{coocc}}=0.08$; the batch marginal $q$ is formed from the reader logits $\hat{s}^{\mathrm{base}}$. Mini-batches are drawn uniformly without class-balanced resampling, so the sampler does not distort the batch label distribution. Here $q$ is the model's predicted label marginal on the batch, and the regularizer penalizes any $q$ whose entropy-regularized transport to $b$ incurs high cost under the co-occurrence geometry. The color pathway independently minimizes $\mathcal{L}_{\mathrm{color}}=\mathcal{L}_{\mathrm{WBCE}}(y,\hat{s}^{\mathrm{color}})$ over the color descriptors.

At inference the two logits are fused per sign,
\begin{equation}
\hat{s}^{\mathrm{final}}_c=\hat{s}^{\mathrm{base}}_c+\gamma_c\,\hat{s}^{\mathrm{color}}_c,\qquad
\hat{y}_c=\mathbf{1}\!\left[\sigma(\hat{s}^{\mathrm{final}}_c)\ge\theta_c\right],
\label{eq:fusion}
\end{equation}
where the per-sign fusion weight $\gamma_c$ and decision threshold $\theta_c$ are selected on validation by grid search rather than by gradient. We set $\gamma_c=0$ for every sign except TonguePale, so the color pathway corrects only the pale-tongue prediction. Because $\gamma_c$ is a fixed validation-selected coefficient, C1 is a color-prototype transport pathway with late fusion, not a learned input-dependent gate.

%% file: sections/04_experiments.tex
\section{Experiments}
\label{sec:experiments}

\subsection{Dataset, Protocol, and Implementation}
\label{sec:dataset}

TongueDx~\cite{bib:drtongue} contains $5{,}109$ images from $4{,}650$ subjects, split into $3{,}371$ training, $843$ validation, and $895$ test images, each annotated with eight non-exclusive binary signs. The split, feature cache, and pseudo-mask grid are fixed and shared across all DINOv2 readout experiments, so every comparison is on identical data and features. Although the split is constructed at the image level, we verified subject-level disjointness across all three splits. The numbers of overlapping subjects were $0$ for train--validation, $0$ for train--test, and $0$ for validation--test. The two rarest signs, TonguePale and Ecchymosis, have only $93$ and $82$ test positives, respectively, and therefore contribute disproportionately to macro-F1 variability.

The primary metric is validation-thresholded test macro-F1: per-class F1-optimal thresholds are searched on the validation split (grid $0.05$--$0.95$, step $0.01$) and applied unchanged to the test split; we also report macro-AUROC, micro-F1, and weighted-F1. All models are trained with the weighted BCE of Section~\ref{sec:objective}. The frozen DINOv2-Large backbone encodes each image at $518\times518$ into a $37\times37\times1024$ patch grid, cached in fp16. Pseudo tongue masks are produced once by a frozen SAM-family segmentation model~\cite{bib:sam,bib:medsam} prompted by a vision-language tongue detector and applied identically to the three splits, with no manual segmentation. The reader is trained with AdamW (lr $3\times10^{-4}$, weight decay $10^{-4}$), batch size $64$, dropout $0.2$, for up to $120$ epochs with early stopping (patience $15$) on validation macro-F1. Color features are eight per-patch LAB/HSV statistics on the same grid, computed on the $518\times518$ bilinearly resized image with OpenCV COLOR\_RGB2LAB and COLOR\_RGB2HSV, normalized as $L/255$, $(a{-}128)/128$, $(b{-}128)/128$ (zero-centered), $H/179$, and $S,V/255$, then averaged per patch with the per-patch standard deviations of $L$ and $S$.

Every per-class threshold and color fusion weight is selected on validation. Single-seed numbers use each model's own validation-selected threshold and color weight. For each seed, the Pale fusion weight is selected from $\{0,0.5,1,1.5,2,2.5,3\}$ on validation macro-F1. The deployed ensemble averages the logits of the six seeds $\{42,\dots,47\}$, each first fused with that seed's validation-selected color weight, and selects its decision thresholds on the ensemble validation logits. The deployed SEAT configuration is then evaluated on test once, with no hyperparameter selected using test performance.

\subsection{Validation-Consistent Ensemble Result}

Under our validation-only operating-point selection protocol, the deployed SEAT ensemble attains $74.15$ test macro-F1 on TongueDx (Table~\ref{tab:main}). The color pathway lifts the same ensemble from its no-C1 value of $73.32$, with TonguePale F1 rising from $41.4$ to $48.1$; the ensemble also attains macro-AUROC $88.10$, micro-F1 $84.89$, and weighted-F1 $85.14$. SEAT's gains concentrate on the rare, cue-defined signs that motivate the design: it is strongest on TonguePale and Ecchymosis, while remaining competitive with strong baselines on easier high-prevalence signs. The seed-42 controlled run reaches $74.67$, its strongest single seed (Table~\ref{tab:main}).

\begin{table*}[t]
\caption{Per-class test F1 on TongueDx (\%). DINOv2-L (global), ResNet-34, Qwen3-VL-4B LoRA, ShizhenGPT-7B-VL, and SEAT use the fixed image-level split with per-class thresholds selected on validation; both vision-language baselines are LoRA-tuned. SEAT is reported both as a single model (mean $\pm$ std over seeds $42$--$47$; per-class columns show the mean, only the Macro column carries $\pm$std) and as the deployed six-seed ensemble. ResNet-50 and Dr.\ Tongue values are reproduced from Table~3 of~\cite{bib:drtongue} as reported contextual references and are not directly comparable to our protocol. Bold marks the highest value across all rows.}
\label{tab:main}
\centering
\footnotesize
\setlength{\tabcolsep}{5pt}
\begin{tabular}{lccccccccc}
\hline
Method & Pale & TipSide & Spot & Ecchy. & Crack & Tooth. & FurTh. & FurYe. & Macro\\
\hline
DINOv2-L (global) & 35.04 & 61.56 & 73.77 & 31.29 & 91.21 & 81.04 & 98.58 & 56.19 & 66.09\\
ResNet-34 & 39.85 & 64.12 & 80.80 & 16.15 & 92.24 & 82.82 & 98.64 & 74.67 & 68.66\\
ResNet-50~\cite{bib:drtongue} & 32.68 & 66.19 & 76.44 & 24.72 & 91.17 & 84.31 & \textbf{98.65} & 76.27 & 68.80\\
Qwen3-VL-4B LoRA~\cite{bib:qwen3vl} & 36.73 & 59.59 & 77.10 & 27.72 & 91.84 & 76.10 & 98.58 & \textbf{78.48} & 68.27\\
ShizhenGPT-7B-VL~\cite{bib:shizhengpt} & 38.83 & 64.97 & 80.59 & 36.51 & 91.26 & 81.34 & 98.58 & 77.17 & 71.16\\
\begin{tabular}[c]{@{}l@{}}Dr.\ Tongue~\cite{bib:drtongue}\\(reported; not directly comparable)\end{tabular} & 46.00 & \textbf{68.22} & \textbf{81.09} & 38.02 & 91.90 & 82.82 & 98.54 & 76.41 & 72.87\\
SEAT (single, mean$\pm$std) & 43.78 & 67.20 & 80.04 & 37.33 & \textbf{92.56} & \textbf{84.86} & 98.58 & 76.33 & $72.59{\pm}1.04$\\
SEAT (ensemble) & \textbf{48.09} & 67.01 & 80.00 & \textbf{44.66} & 92.48 & 84.26 & 98.58 & 78.15 & \textbf{74.15}\\
\hline
\end{tabular}
\end{table*}

\subsection{Multi-Seed Component Effects}
\label{sec:perseed}

With all selection on validation, we report the full $2\times2$ factorial of the two components across seeds $42$--$47$ (Table~\ref{tab:seeds}). A test-image bootstrap corroborates the paired gains: the $95\%$ CIs for C1 over the reader ($[+0.21,+0.78]$), C2 over the reader ($[+0.09,+0.79]$), and the full model over the reader ($[+0.61,+1.61]$) all lie above zero. C1 is the primary component, improving all six seeds both alone (Reader\,$\to$\,+C1, mean $+0.50$, six of six) and with C2 (+C2\,$\to$\,+C1+C2, mean $+0.68$, six of six). C2 is secondary and complementary: modest alone (Reader\,$\to$\,+C2, mean $+0.44$, four of six), but in the deployed model, which always includes C1, it adds $+0.61$ on five of six seeds, with a small positive interaction ($+0.18$), indicating that the two components are complementary rather than redundant. The one exception is seed~$45$, where C1 alone ($72.50$) already exceeds the full model ($71.46$), so C2 costs $-1.04$ on top of C1 there; it helps on the other five seeds. Beyond accuracy, C2 supplies the inspectable label-structure evidence type used in the audit (Section~\ref{sec:audit}).

\begin{table}[t]
\caption{Component $2\times2$ factorial (seeds 42--47), all selection on validation. Top: validation-thresholded test macro-F1 as mean $\pm$ population std across seeds, with a test-image bootstrap $95\%$ CI. Bottom: paired gains with their bootstrap CIs. C1 improves every seed both alone and with C2; C2 is secondary and complementary (see text).}
\label{tab:seeds}
\centering
\footnotesize
\setlength{\tabcolsep}{4pt}
\begin{tabular}{lcc}
\hline
Configuration & Macro-F1 & Bootstrap $95\%$ CI\\
\hline
Reader & $71.47 \pm 0.93$ & $[70.43,\,72.52]$\\
Reader\,+\,C1 & $71.97 \pm 0.89$ & $[70.93,\,73.04]$\\
Reader\,+\,C2 & $71.91 \pm 1.24$ & $[70.86,\,72.95]$\\
Reader\,+\,C1\,+\,C2 & $\mathbf{72.59 \pm 1.04}$ & $[71.53,\,73.67]$\\
\hline
Paired gain & Gain & Bootstrap $95\%$ CI\\
\hline
C1\,$-$\,Reader & $+0.50$ & $[+0.21,\,+0.78]$\\
C2\,$-$\,Reader & $+0.44$ & $[+0.09,\,+0.79]$\\
Full\,$-$\,Reader & $+1.11$ & $[+0.61,\,+1.61]$\\
Full\,$-$\,(Reader\,+\,C1) & $+0.61$ & $[+0.23,\,+1.00]$\\
Full\,$-$\,(Reader\,+\,C2) & $+0.68$ & $[+0.29,\,+1.05]$\\
\hline
\end{tabular}
\end{table}

\subsection{Single-Seed Controlled Decomposition}

We isolate each component at seed $42$ on the same $8$-head reader. C2 raises macro-F1 from $73.42$ to $74.37$ (micro- and weighted-F1 up, macro-AUROC flat at $86.9$), and C1 at validation-selected $\gamma{=}1.0$ then raises it to $74.67$, improving all four metrics (macro-F1 $74.67$, macro-AUROC $87.29$, micro-F1 $85.01$, weighted-F1 $85.67$).

\begin{figure}[t]
\centering
\includegraphics[width=\columnwidth]{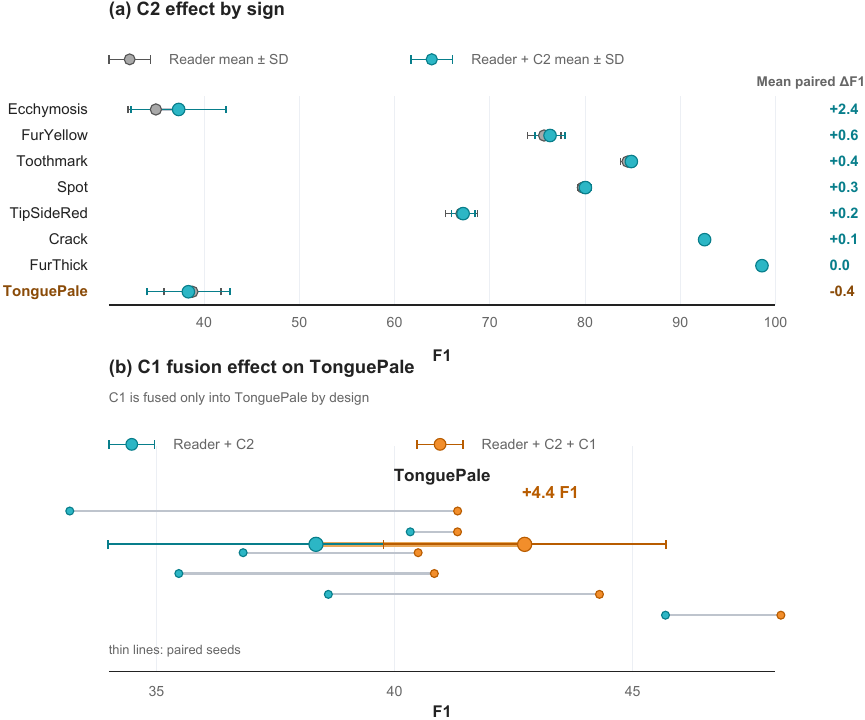}
\caption{Multi-seed component effects across seeds 42--47.
(a) Per-sign F1 for the Reader and Reader+C2, reported as
mean $\pm$ standard deviation across seeds; labels show the
mean paired change. (b) TonguePale F1 before and after C1
fusion on top of Reader+C2; thin lines connect matched seeds.
All thresholds and fusion weights are selected on validation.
Aggregate macro-F1 results and bootstrap confidence intervals
are reported in Table~\ref{tab:seeds}.}
\label{fig:perclass}
\end{figure}

Per class (Fig.~\ref{fig:perclass}), C2 improves five of eight signs (largest on Ecchymosis $+5.3$, TonguePale $+1.6$) and C1 adds $+2.4$ to TonguePale by construction.

\subsection{Supporting Analyses}

\begin{figure*}[!t]
\centering
\includegraphics[width=0.68\textwidth]{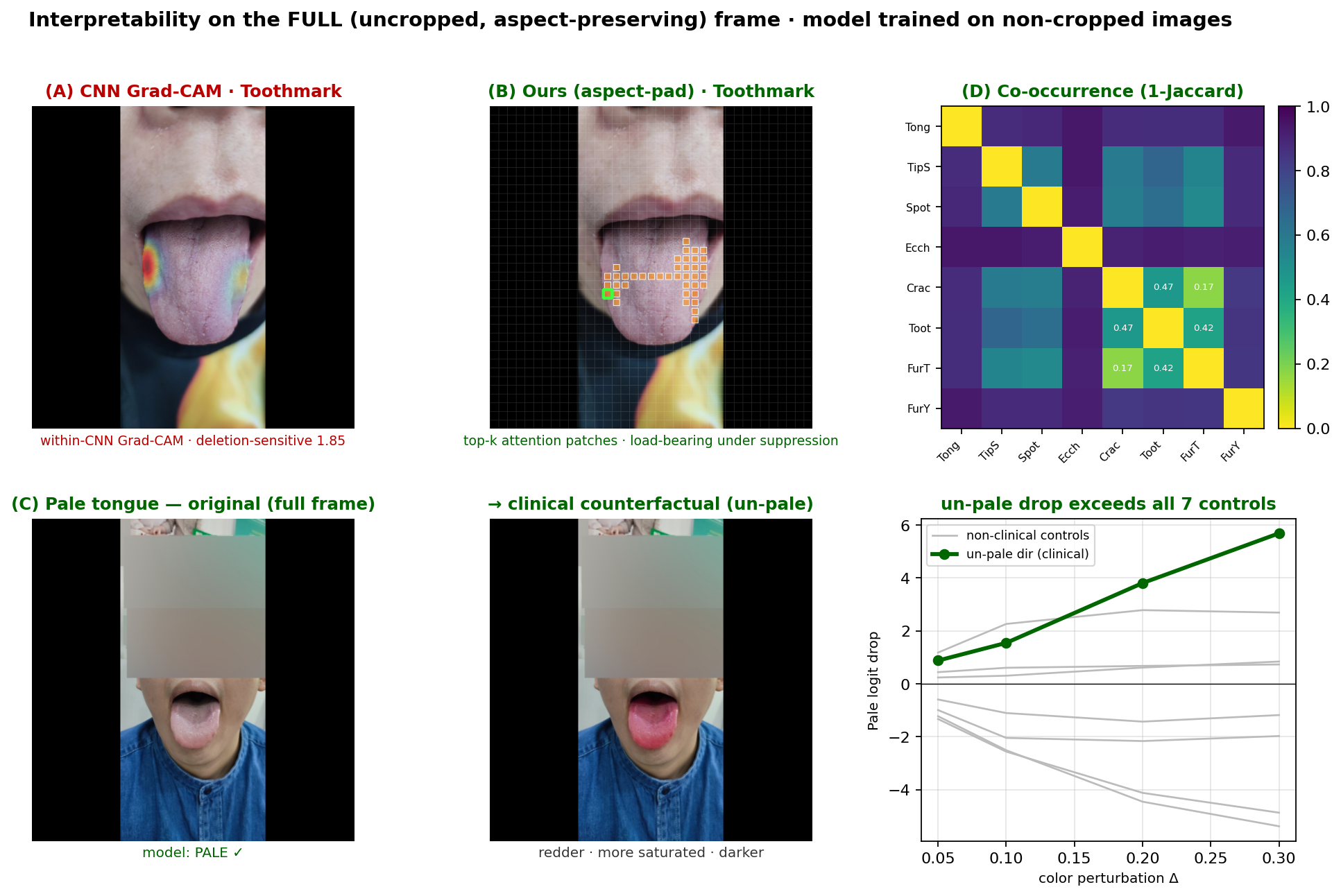}
\caption{Evidence-type coverage, read as a capability matrix. Left to right: (A) a ResNet-34 Grad-CAM, deletion-sensitive within the CNN at $1.851$ on Toothmark; (B) our attention-based spatial evidence, load-bearing under readout-mask suppression ($2.386$, across Toothmark, Crack, and Spot); (C) a descriptor-space color intervention on the color pathway for the pale-tongue sign, whose directed effect exceeds every one of seven matched non-clinical color directions and which a spatial map does not express; and (D) the sign co-occurrence geometry ($1-\text{Jaccard}$) used by C2, an output-space structure rather than an image map.}
\label{fig:interp}
\end{figure*}

Sign-type evidence audit.
\label{sec:audit}
We evaluate each prediction with a procedure matched to its evidence type: an intervention test for spatial and chromatic cues, and structural inspection for label-level co-occurrence structure.

For localized morphological signs (Toothmark, Crack, Spot) we use a readout-mask suppression test. Given a patch importance score from the reader, we set the mask weights of the top-ranked patches to zero at increasing counts, re-run the reader, and measure the drop in the target logit; the deletion-AUC is the normalized area under this drop curve. Because the stabilizing constant $\delta$ retains finite softmax support and the backbone is not re-run, this test strongly suppresses rather than removes patch access, and measures readout reliance rather than pixel-level removal. Within the seed-42 OT-readout runs, suppressing the patches selected by the attention map is load-bearing for these signs (mean AUC $2.386$ in the SignOT run vs.\ $0.003$ for random patches), whereas the OT selector is load-bearing only when it effectively copies attention and is otherwise diffuse (AUC $\approx0$). Because the deployed reader uses the same mask-guided attention mechanism, this diagnostic supports the use of attention-based spatial evidence. As a within-CNN reference, a ResNet-34 Grad-CAM shows comparable sensitivity on Toothmark, $1.851$ versus random $0.759$; this is a single sign under pixel occlusion on a different model and so not directly comparable.

For the pale-tongue sign, patch-level suppression is not an appropriate test because the concept is not a small region. We instead apply a clinically motivated descriptor-space intervention to the color pathway: within the tongue mask we move the cached per-patch LAB/HSV descriptors away from paleness along the clinical axis (decreasing $L$, increasing $a$ and $S$) by a magnitude $\Delta\in\{0.05,0.1,0.2,0.3\}$ on the normalized channel scale, re-evaluate the color pathway's Pale logit without re-running the DINOv2 reader, and compare against the seven matched non-clinical controls, each applying the same per-channel magnitude with one of the other sign combinations on the same three channels. This measures the sensitivity of the color pathway whose output forms the fused Pale correction. Reporting the mean Pale-logit drop over Pale-positive test images, the counterfactual-AUC integrates this mean-drop curve over $\Delta$. The directed intervention gives $3.215$ and exceeds every one of the seven non-clinical controls, whose mean is $-0.97\pm2.13$, a margin of $4.19$ AUC units with image-level $95\%$ bootstrap CI $[4.06,4.31]$; the directed drop rises monotonically from $0.88$ at $\Delta{=}0.05$ to $5.69$ at $\Delta{=}0.3$. This indicates that the auxiliary color pathway responds preferentially to the paleness direction rather than to color change in general; it does not by itself establish that the deployed ensemble's Pale decision relies causally on this pathway. For label-level co-occurrence structure, C2 yields an inspectable output-space geometry that we read directly from the induced transport coupling and its $1-\text{Jaccard}$ cost. The point of the audit is coverage (Fig.~\ref{fig:interp}): a single spatial saliency map does not express chromatic sensitivity or output-level label structure.

Not every rare class benefits from an added low-level modality. Pale separates strongly on color ($a\!=\!-0.73\sigma$, $S\!=\!-0.67\sigma$, $L\!=\!+0.50\sigma$) and gains from color fusion. Ecchymosis separates by $<0.25\sigma$ on every color/texture channel; color-only and texture-only detectors, at $19.7$ and $20.0$ F1, are far below the DINOv2 semantic prediction. Under the examined color and texture probes we did not find a complementary low-level signal for Ecchymosis, and its validation-selected color-fusion weight was zero, so the color pathway is restricted to Pale.

\subsection{Exploratory and Negative Findings}
\label{sec:exp-ablation}

All ablations in this subsection are run at seed 42 on identical features. Patch-side OT readouts underperform. We extensively evaluated OT-based feature aggregation~\cite{bib:otke} as a discriminative readout across more than twelve formulations: additive prototype-to-patch OT (SignOT), attention-guided sparse/unbalanced/partial OT (AOT), and OT-generalized attention that provably contains attention at $\lambda\!=\!0$ (OTGen). Table~\ref{tab:otreadout} summarizes representative configurations; none beats the attention reader's $73.42$, and all fall within $70.81$--$72.95$. A learned interpolation converges near the attention limit ($\lambda\!\approx\!0.02$), further indicating that OT does not improve the discriminative patch reader; OT appears more useful as a color- or label-side inductive bias than as a readout replacement.

\begin{table}[t]
\caption{OT-as-readout ablation at seed 42 (validation-thresholded macro-F1). All variants $<$ attention reader.}
\label{tab:otreadout}
\centering
\begin{tabular}{lc}
\hline
Configuration & Macro-F1\\
\hline
Attention reader & \textbf{73.42}\\
SignOT: global + OT (joint / frozen) & 72.95 / 72.76\\
SignOT: OT only (no global) & 70.81\\
AOT: attention-demand sparse (balanced) & 72.54\\
AOT: unbalanced $\rho{=}0.1$ / partial $m{=}0.7$ & 72.18 / 72.49\\
OTGen: $K{=}4$ mean / $K{=}4$ concat & 72.45 / 72.58\\
\hline
\end{tabular}
\end{table}

\begin{table}[!t]
\caption{Color-pathway ablation at seed 42 (validation-thresholded test F1). C1's prototype OT is a mild refinement over softmax-prototype attention; only TonguePale is fused into the deployed model.}
\label{tab:colorbase}
\centering
\footnotesize
\setlength{\tabcolsep}{3.5pt}
\begin{tabular}{lcccc}
\hline
Color model & Macro-F1 & AUROC & Micro-F1 & Pale-F1\\
\hline
global LAB/HSV\,+\,MLP & 59.53 & 61.53 & 71.88 & 27.78\\
softmax-prototype & 63.74 & 69.74 & 76.95 & 37.89\\
C1: prototype OT & \textbf{64.79} & \textbf{71.65} & \textbf{78.28} & \textbf{40.15}\\
\hline
\end{tabular}
\end{table}

An explicit color pathway helps, but OT is a mild refinement. We compare three color-only detectors on the same seed-42 descriptors (Table~\ref{tab:colorbase}). An explicit, structured color pathway is necessary: a global mask-weighted LAB/HSV MLP is far weaker on every metric. Within the color-prototype family, the OT formulation gives only a small edge over plain softmax-prototype attention, of which it is a strict generalization: softmax-prototype attention is the $\lambda_{\mathrm{col}}\!\to\!0$ limit of Eq.~\eqref{eq:colorot}. We therefore present OT as a mild refinement rather than an essential ingredient. On TonguePale, the only sign fused into the deployed model, the ordering is clean: prototype OT $40.15 >$ softmax-prototype $37.89 >$ global MLP $27.78$.

The marginal formulation is preferred. A finer per-pair co-occurrence constraint reached only $73.92$ macro-F1, below the marginal co-occurrence regularizer's $74.37$, so we keep the marginal formulation.

Late fusion outperforms the integrated dual-branch model. Folding the color and co-occurrence terms into a single end-to-end dual-branch OT model scores $72.68$, below late fusion's $74.67$, because its OT-generalized DINOv2 branch is weaker than the tuned attention reader. Retaining the tuned attention reader and adding color through late fusion is preferable to the integrated dual-branch alternative.

\subsection{Limitations}
\label{sec:limitations}

Our claims are bounded on three fronts. Statistical scope: the rare-class test sets are tiny (TonguePale $93$, Ecchymosis $82$ positives) and single-model macro-F1 varies by about $3.6$ points across seeds, which is why we report the color effect as a per-seed paired gain and deploy a validation-selected ensemble. Data scope: results are on a single dataset with pseudo tongue masks; although the test set is subject-disjoint from training (Section~\ref{sec:dataset}), we did not evaluate cross-site or cross-device generalization. Design scope: the backbone is frozen rather than fine-tuned, the chromatic evidence test is a descriptor-space rather than image-space intervention, and combining both OT gains in one end-to-end model without weakening the reader remains future work. The suppression audit is run on seed-42 readouts rather than on every deployed ensemble member, and the label co-occurrence geometry is an inspectable structure rather than independently verified relation evidence. The color descriptor also averages hue linearly, pools all patch pixels, and applies no illumination normalization, so hue wrap-around, boundary contamination, white balance, and device variation may affect C1.

%% file: sections/06_conclusion.tex
\section{Conclusion}

We presented SEAT, which treats the type of evidence, rather than the individual
label, as the unit of design for multi-label tongue-sign recognition. On a frozen
DINOv2 backbone, SEAT pairs each clinically distinct cue with a matching
prediction mechanism and, where applicable, an evidence audit: a
per-sign attention reader for localized morphology, a dedicated color pathway for
the chromatic pale-tongue sign, and a co-occurrence-informed regularizer for
output-space label structure. On TongueDx, a six-model ensemble reaches $74.15$
macro-F1, surpassing both a strong same-backbone baseline and a substantially
larger domain-specialized vision-language model, with the gains concentrated on
the rare, cue-defined signs the design targets. Beyond accuracy, this reframes
interpretability for clinical decision support from where a model looks to
whether each prediction responds to the evidence its clinical definition calls
for. The study is limited to a single dataset with pseudo masks and found no
complementary low-level signal for one rare sign; integrated multi-pathway
training, gold-standard masks, and cross-site validation remain future work.